%% file: root.tex
\documentclass[letterpaper, 10 pt, conference]{ieeeconf}  

\IEEEoverridecommandlockouts                              

\usepackage{cite}
\usepackage{amsmath,amssymb,amsfonts}
\usepackage{algorithmic}
\usepackage{graphicx}
\usepackage{sidecap}
\usepackage{textcomp}
\usepackage{lipsum}  
\usepackage{xcolor}
\usepackage[font=footnotesize]{caption}
\usepackage{setspace}
\usepackage{tabularx, booktabs}
\usepackage{subcaption}
\usepackage{censor}
\usepackage{hyperref}
\usepackage{svg}

\hypersetup{
    colorlinks=true,
    linkcolor=blue,
    filecolor=magenta,      
    urlcolor=cyan,
    pdftitle={Overleaf Example},
    pdfpagemode=FullScreen,
    }

\title{\LARGE \bf
How Well Do Pseudo-Rigid Body Models Capture Real Plants? In-Field Validation of Simulated Blueberry Canes
}

\author{Hannah Kolano$^{1}$, Chelse VanAtter$^{1}$, Wei Yang$^{2}$, Cindy Grimm$^{1}$, Joseph R. Davidson$^{1}$
\thanks{This work was supported by the AI Research Institutes program supported by NSF and USDA-NIFA under the AI Institute: Agricultural AI for Transforming Workforce and Decision Support (AgAID) Award No. 2021-67021-35344.}%
\thanks{$^{1}$Collaborative Robotics and Intelligent Systems (CoRIS) Institute, Oregon State University, Corvallis OR 97331, USA {\tt\footnotesize \{kolanoh,vanattec,cindy.grimm, joseph.davidson\}@oregonstate.edu}}
\thanks{$^{2}$Department of Horticulture, Oregon State University, Corvallis, OR 97331, USA, {\tt\footnotesize wei.yang@oregonstate.edu}}
}

\begin{document}

\maketitle
\thispagestyle{empty}
\pagestyle{empty}

\begin{abstract}
Many labor-intensive tasks in fruit production such as pruning require physically interacting with the plant (e.g., pushing, pulling, bending limbs, etc.). Due to increasing labor shortages, there is widespread interest in the adoption of robotics in this area. When the robot must physically interact with the system, a deformable model of the plant is beneficial. Coupled, spring-loaded beams have been used in prior work to build deformable plant models in simulation for learning, planning, and control, but this approach has rarely been validated against the mechanical properties of real-world plants. In this paper, we use this rigid-body model, grounded in real material properties and beam bending theory, to simulate the deformation of blueberry canes under loading. To validate the model, we simulate the canes in MuJoCo and compare their behavior with real-world data collected from probing canes at a commercial farm with a custom testbed. We perform sensitivity analysis on several key modeling variables and show that this approach is highly sensitive to the plant diameter calculations and a priori flexural modulus estimation, which is dependent on season and blueberry variety. We also share our dataset of live blueberry cane deformation, including RGB-D images and measured forces and displacements.  

\end{abstract}

\newcounter{mycounter}
\setcounter{mycounter}{1}


\input{Sections/01_Intro}

\input{Sections/02_Background}

\input{Sections/03_Methods}

\input{Sections/04_Experiments}

\input{Sections/05_Results}

\input{Sections/06_Conclusion}

\bibliographystyle{./bibliography/IEEEtran}
\bibliography{./bibliography/BlueberryDyn.bib}

\end{document}

%% file: Sections/01_Intro.tex
\section{Introduction}


Rising production costs and increasing labor shortages motivate the development of robotic cultivation of high-value, perennial specialty crops (e.g., grapes, tree fruit, berries). Many in-field management activities  such as plant training, pruning~\cite{You_2023,Silwal_2022,WILLIAMS_2023}, thinning~\cite{Bhattarai_2024,Williams_2024}, and harvesting~\cite{Rajendran_2024} involve physically interacting (e.g., pushing, pulling, bending, tying, etc.) with the plant in complex, unstructured settings. These types of dexterous manipulation tasks require models of the environment to guide robot planning and control.  


A typical model of woody bushes, trees, etc. considers all branches to be completely rigid and fixed, and thus path planning algorithms avoid colliding with the plant's structure during task execution. However, this assumption of rigidity is overly restrictive and prevents the robot from performing any task in which a branch would have to be moved out of the way to aid task completion. 
This type of activity requires treating the plant as deformable instead of rigid. 


Prior work has often focused on visually realistic plant modeling \cite{quigley_real-time_2018, Aubry_Xian_2015}, or on whole-tree vibration models for plants under wind conditions \cite{james_mechanical_2006, Pivato_Dupont_Brunet_2014}. While there has been some recent work on using mass-spring models in simulation \cite{jacob_learning_2024, Munn_2026}, these models are frequently not tested against real plants to evaluate simulation behavior. 

\begin{figure}
    \centering
    \includegraphics[width=1.0\linewidth]{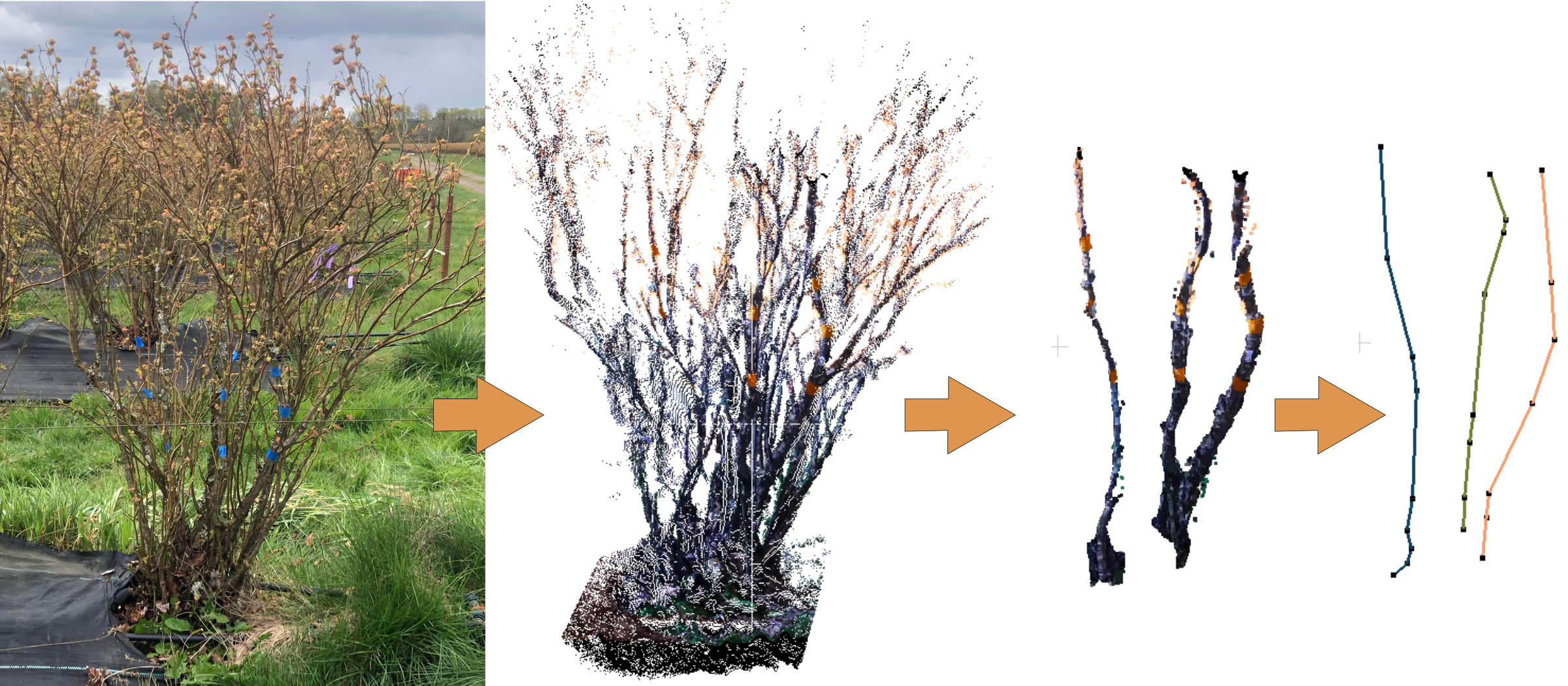}
    \caption{Workflow adopted in this study. Blueberry plants (\textit{left}) are imaged during the dormant season. Individual canes are then extracted from the point cloud and discretized into serial chains of links with compliant joints (\textit{right}). These chains are probed in a physics simulator, with the predicted behavior compared with in-field measurements from a commercial farm.}
    \label{fig:abstract}
    \vspace{-10pt}
\end{figure}


We posit that physics simulations for robotic plant manipulation should accurately represent force-displacement behavior -- they should output realistic forces when used in sim-to-real applications. In our study, we probe a blueberry bush in real-world conditions, a typical pruning task behavior, to measure actual force-displacement behavior in the field and compare it to simulation outputs.  

We first collected the test data by probing commercial blueberry plants with a custom testbed. We built digital models of 18 unique canes by discretizing point clouds of blueberry bushes into serial chains of rigid bodies (Fig.~\ref{fig:abstract}) and used real material properties and Euler-Bernoulli beam theory to select joint compliance. To evaluate the lumped-parameter model, we compared force-deflection data from probing the simulated plant models in MuJoCo \cite{mujoco} with the collected field data. Additionally, we performed sensitivity analysis on several modeling variables to identify potential significant sources of error. 

In summary, the contributions of this paper are:
\begin{itemize}
\item A dataset capturing live blueberry cane deformation (visual deformation under load, plus measured forces and displacements)\footnote{Data can be accessed at \url{https://figshare.com/s/f82e41c7115187e42df8}}.
\item Evaluation of the physics model against real-world data collected from a commercial blueberry farm with a custom testbed.
\item Sensitivity analysis of several variables for lumped-parameter rigid-body models.
\end{itemize}



This paper explores the impact of season, age, and variety on the flexural modulus of blueberry plants, and its impact on the error of our simulation. We also explore the impact of full-cane diameter estimation. Our na\"{i}ve sim-to-real comparison shows significant error, even for a simple, no-branching case, but we show that the lumped parameter model is viable for simple canes, subject to natural variations that should be accounted for in modeling practices. 


The rest of this paper is organized as follows. Section~\ref{sec:background} reviews the prior work on tree modeling and robotic branch manipulation. Our testbed, in-field data collection procedure, and deformable plant model are described in Sec.~\ref{sec:Methods}. We next describe our method of obtaining flexural modululus measurements.  Section~\ref{sec:Experiments} summarizes the physics simulation experiments in MuJoCo. We compare predicted stiffness and force/deflection profiles between the simulated canes and in-field measurements in Sec.~\ref{sec:Results}, as well as discuss the results of the sensitivity analysis. Finally, in Sec.~\ref{sec:Conclusion} we summarize our findings and their implications for future work.






%% file: Sections/02_Background.tex
\section{Background}
\label{sec:background}

\subsection{Mechanical models of plants}
Trees are commonly modeled as a cascading system of oscillating components, starting with the trunk that bends as a clamped beam, and branches that bend independently~\cite{de_langre_plant_2019, sellier_finite_2006}. 
Pivato et al. use a single discretized trunk and Euler-Bernoulli beam theory to perform modal analysis on coniferous trees~\cite{Pivato_Dupont_Brunet_2014}. James et al. posited a dynamic structural model of trees comprised of spring-mass-dampers, and calculate the Young's modulus based on that topology~\cite{james_mechanical_2006}. 

Other attempts to obtain the Young's modulus of live trees include Kawaguchi et. al, who attempted a non-destructive measurement method on live Eucalyptus trees~\cite{KAWAGUCHI_IMAEDA_ZHANG_MUTO_2024}. Importantly, they found that live trees have large sample-by-sample variation in their material properties, and therefore refrained from reporting an average value. A study by Cannell and Morgan found the Young's modulus for both trunks and branches of coniferous trees~\cite{Cannell_Morgan_1987}. They found relationships between the Young's modulus and the specific gravity of the sample, as well as its water content. 

\subsection{Robotic applications with deformable plant models}
Whereas simplified versions of Featherstone's rigid-body dynamics algorithms have been developed to generate animations for tree graphics~\cite{quigley_real-time_2018}, there are few models that exist specifically for autonomous deformable object manipulation in agricultural applications. Rijal et al.~\cite{rijal_force_2025} proposed a modified RRT* path planner that incorporates branch geometric constraints based on a branch heuristic model; force feedback is used during online re-planning to protect branches from damage. Parayil et al.~\cite{Parayil_Peynot_Lehnert_2025} also use force feedback to prevent plant damage without a full dynamic model. Katyara et al.~\cite{Katyara_Ficuciello_Caldwell_Chen_Siciliano_2021} generated realistic grape vines in Blender to simulate pruning tasks with an interaction controller, but only simulate local dynamics of the vines. Kim et al.~\cite{Kim_2024} represent the state of a tree as a graph using mass-spring-damper models. Graph neural networks were used to learn a contact policy for inferring actions to manipulate trees toward target states. While the method was evaluated in simulation, validation in the real world was left for future work. 

Yandun et al.~\cite{Yandun_Silwal_Kantor_2020} use Quigley's~\cite{quigley_real-time_2018} simplified mass-spring dynamics formulation to predict deformation of a grape vine in a laboratory setting, but use their own force-displacement data to find the relevant stiffness values, making their predictions circular. Jacob et al.~\cite{jacob_learning_2024} recently developed artificial trees in ISAAC Gym reminiscent of James' dynamic model~\cite{james_mechanical_2006}. They then trained an inference algorithm based on Stein Variational Gradient Descent (SVGD) to infer spring parameters from reference trajectories obtained by active probing of the tree. Most recently, Munn used a nearly identical lumped-parameter model to build a simulated orchard environment for apple-picking \cite{Munn_2026} -- and specifically points out validation as an important next step. We build on these earlier works with an approach that uses real plant material properties, estimates plant mechanics without the need for active probing, and, most notably, compares plant simulations with equivalent real-world measurements from a commercial farm.





%% file: Sections/03_Methods.tex
\section{Methods}
\label{sec:Methods}

\subsection{In-field data collection}
We collected in-field experimental data at a commercial blueberry farm (`Legacy' blueberry variety) in Spring 2024. This data included RGB-D images and force/deflection measurements from physically pushing on the canes with an automated probe. 

\begin{figure}
    \centering
    \vspace{5pt}
    \includegraphics[width=1.0\linewidth]{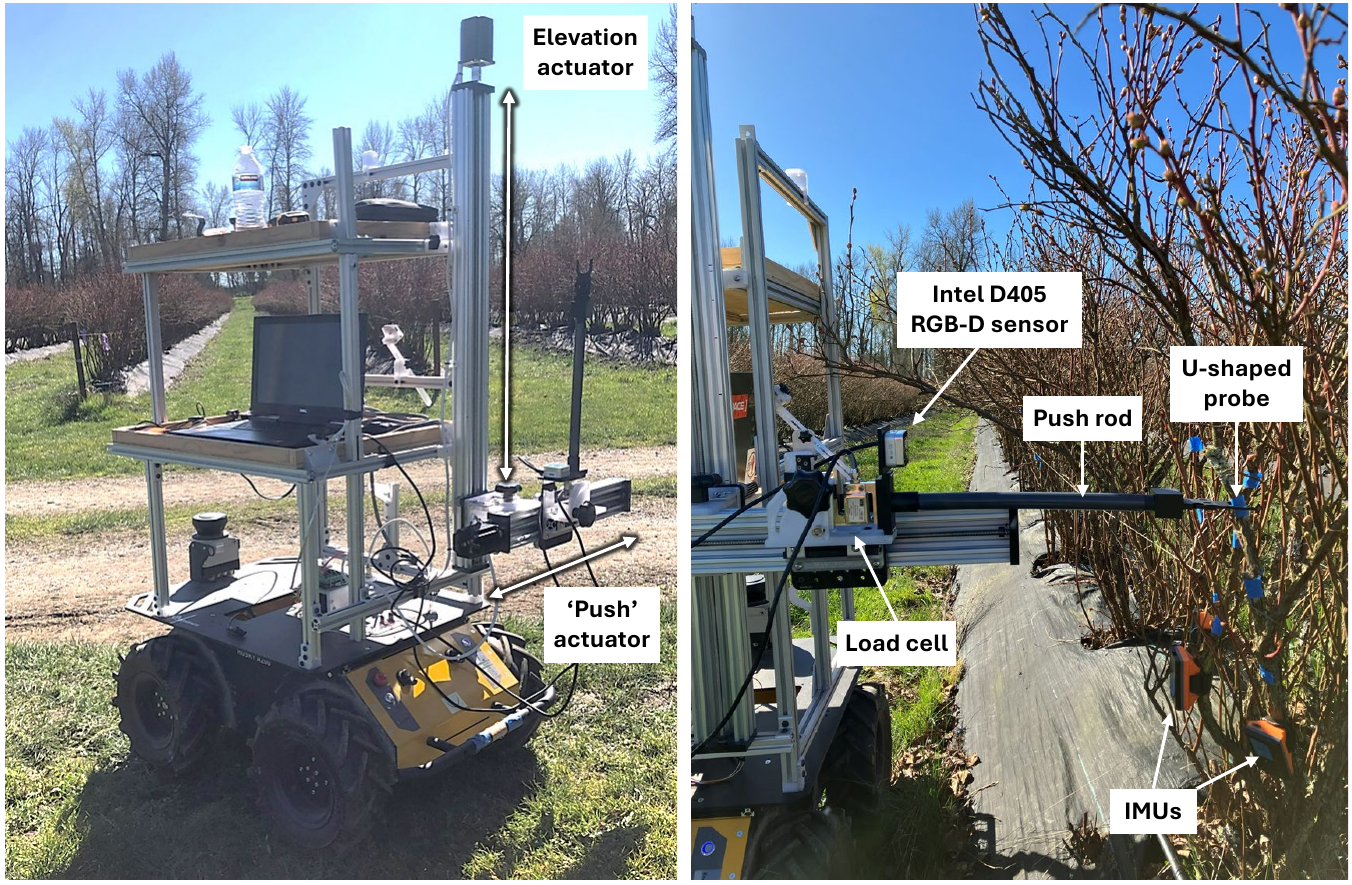}
    \caption{\textit{Left:} Mobile robot with custom 2 degree-of-freedom cartesian gantry system. \textit{Right:} Up-close view of components during linear probing of a blueberry cane. Two IMUs installed on the plant are shown at the bottom right of the image.}
    \label{fig:pushingapparatus}
    \vspace{-10pt}
\end{figure}

\subsubsection{Hardware}
The measurement testbed, shown in Fig.~\ref{fig:pushingapparatus}, is a portable 2 degree-of-freedom actuated gantry system built on top of a mobile Clearpath Robotics (Kitchener, Canada) Husky A300. A NEMA 23 stepper motor on a lead screw actuates the entire carriage up and down to position the unit at the correct elevation. A second NEMA 23 stepper motor with lead screw actuates a plastic pushing arm forward/backward to exert force on the cane with a U-shaped probe. The pushing arm is mounted to a single axis load cell with an amplifier that measures compressive force. Passive, lockable joints can be used to manually adjust the yaw and pitch of the pushing arm, which is required to establish an initial perpendicular probe orientation on uneven terrain. The testbed is powered by a battery and controlled with an Arduino Mega microcontroller via ROS commands.  

\subsubsection{Plant imaging}
We imaged the plants with two RGB-D cameras. An Intel (Santa Clara, CA, USA) RealSense D405 camera mounted to the push rod provided an up-close view of the cane during probing. For a better perspective of the entire bush, we also imaged the six plants with a Microsoft (Redmond, WA, USA) Azure Kinect camera. The Kinect was not installed on the mobile testbed; instead it was mounted on an approximately 1 meter long linear actuator clamped to a table. Video was collected as the actuator scanned a path parallel to the bush (i.e. along the axis of the row). 

\subsubsection{Plant probing procedure}
We selected six unique bushes and three canes from each bush, for a total of 18 canes. Only long, slender canes having no intermediate branching junctions (i.e. the base of the cane splits into two or more limbs) were selected for this study. On each cane, we selected three push points spaced four inches apart, which were marked with blue tape, for a total of 54 push points across all bushes. For each push point, we jogged the horizontal and linear actuators until the U-shaped probe was aligned with the tape on the cane, and manually oriented the pitch angle from horizontal such that the pushing arm and branch were perpendicular. The trial run then consisted of the stepper motor actuating forward at 0.5 mm per second for 90 seconds, or until the force limit was hit (1000 gf). We measured the diameter of the canes at each of the push points with calipers, as well as the distance of the push point above the ground (with a measuring tape). 

A 3D printed housing containing an Arduino Nano 33 BLE (Arduino SA, Chiasso, Switzerland) and a small battery was strapped to the base of each of the three selected canes to collect accelerometer and gyroscope data from the plant during probing of that bush (see Fig.~\ref{fig:pushingapparatus}). These modules communicated sensor data over Bluetooth low energy (BLE) to a central host microcontroller, which published the data as Vector3 ROS topics via serial communication.

\subsection{Cane geometry reconstruction} 

\begin{figure*}
    \centering
    \vspace{5pt}
    \includegraphics[width=\linewidth]{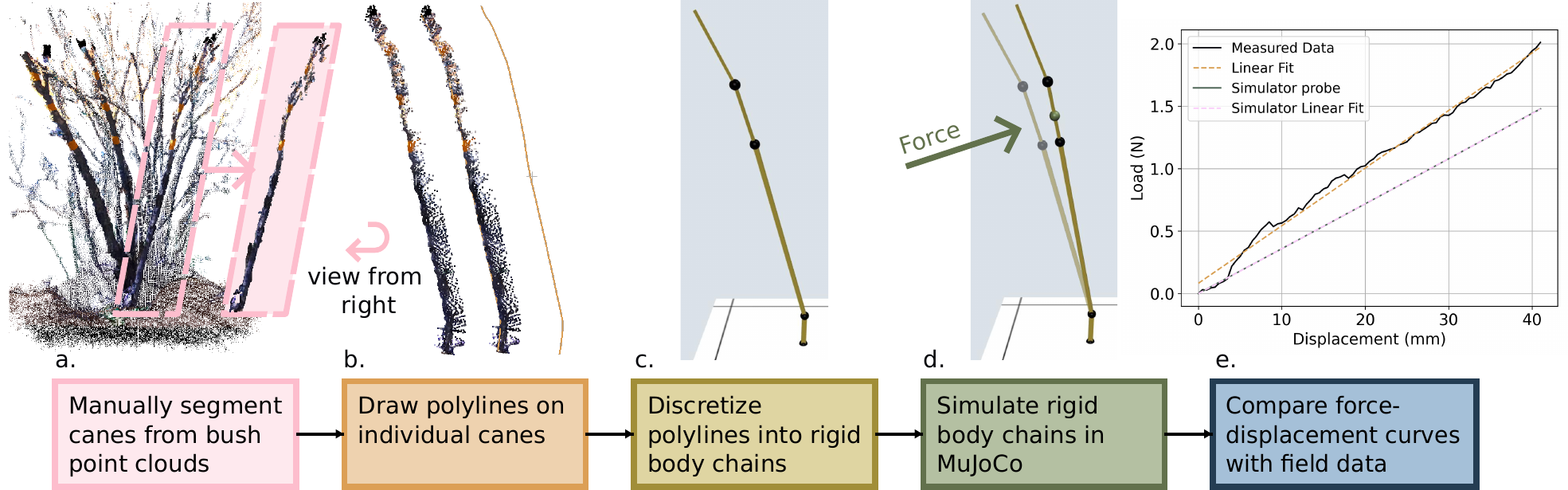}
    \caption{The full testing pipeline. \textit{a.} Three test canes were manually segmented from the blueberry bush point cloud. The push points are visible as orange marks. \textit{b.} CloudCompare's polyline tool was used on each cane. \textit{c.} The polyline was discretized into segments until the maximum distance between the segmented chain and the polyline was 5 mm. Black dots represent joints. \textit{d.} MuJoCo was used to simulate the force applied to the cane as a lumped-parameter, rigid body model. The green dot represents the push point. \textit{e.} The force-displacement slopes from the field data and simulation were compared.}
    \label{fig:pipeline}
    \vspace{-13pt}
\end{figure*}

\label{subsec:reconstruction}
We next reconstructed individual canes from the point clouds of complete bushes. First, we used a manual process to draw polylines on the point clouds; then, a greedy method was used to discretize the polyline into a rigid-body system. 

\subsubsection{Manual polyline generation}
\label{sec:polylines}
CloudCompare's~\cite{CloudCompare_2026} Segment tool was used to segment the point cloud into the individual test canes (Fig.~\ref{fig:pipeline}(\textit{a})). We used CloudCompare's polyline tool to fit a polyline to each cane (Fig.~\ref{fig:pipeline}(\textit{b})). Polylines were smoothed with the built-in Chaikin smoothing tool with five iterations using strength 0.35.

\subsubsection{Curve discretization}
\label{sec: discretization}
Since beam curvature plays a significant role in force loading equations, we implemented a discretization strategy that preserved  cane curvature. We began with a single link with endpoints positioned at the ends of the curve. At each iteration 100 uniform points on the curve were queried, and the point furthest from the current link chain was added as another split in the chain. Endpoints were added until the maximum distance was 5 mm. 

\textit{Assumption \#~\themycounter} is that cane diameter decreases linearly with length along the cane. We used the three in-field diameter and height measurements (per cane) to generate a line of best fit. For each link, the length estimate is to the middle of the link, with the diameter calculated from the best-fit line. Figure \ref{fig:pipeline}(\textit{c}) shows a cane after the discretization with its relevant diameters.
\stepcounter{mycounter}

\textit{Assumption \#~\themycounter} is that we are only dealing with single canes with no significant branching junctions. While there are small diameter branches, buds, etc. at the top of the cane where berries develop, their physical effects are not considered here. Future work will explore modeling approaches that incorporate all parts of the plant. 
\stepcounter{mycounter}

\subsection{Rigid-body model with beam deflection}
We modeled the discretized cane in MuJoCo \cite{mujoco} using the link chain from the previous section, with each link preceded by two coincident rotational joints. Axial twisting of the canes is ignored. 
Each joint has a damping and stiffness value. Since robotic plant manipulation would typically be a low-velocity action, we make a quasi-static assumption (\textit{Assumption \#~\themycounter}) and select a damping value of 0.15 (N-m-sec/rad) for all joints based on other tree values reported in the literature~\cite{Pivato_Dupont_Brunet_2014}. Mass is calculated based on a uniform density of 580 kg/m$^3$~\cite{zhang_more_2022}. 
\stepcounter{mycounter}

To determine individual joint stiffnesses, which significantly influence cane behavior in simulation, we adopt Euler-Bernoulli beam theory for cantilevered beam deflections. We assume beams fixed at one end, with a pure moment at the other end. The tip deflection for such a beam is~\cite{Gere_1997}
\begin{equation}\label{eq:beamdeflect}
    v = \frac{ML^2}{2EI}
\end{equation}
where $v$ (m) is the deflection of the beam, $M$ (Nm) is the applied moment, $L$ (m) is the length of the beam, $E$ (MPa) is the Young's modulus, and $I$ (m$^4$) is the second moment of inertia. Hooke's law for for a linearly compliant joint is  
\begin{equation}\label{eq:torque}
    M = K\theta
\end{equation}
where $K$ (N-m/rad) is the stiffness coefficient, and $\theta$ (rad) is the displacement angle from the equilibrium position. Using the small angle assumption (\textit{Assumption \#~\themycounter}), we can use $\theta=\frac{d}{L}$, rearrange, and substitute back into Eqn.~\ref{eq:torque} to obtain 
\stepcounter{mycounter}
\begin{equation}
    v = \frac{ML}{K}
\end{equation}
Substituting the deflection back into Eqn.~\ref{eq:beamdeflect} yields
\begin{equation}
    K = \frac{2EI}{L}
\end{equation}
for the rotational stiffness of the joint. Therefore, when constructing the rigid-body model, the length of the next link and the cross-sectional area of the link are both taken into account when determining the joint stiffness. Section~\ref{subsec:modulustesting} describes how the modulus was selected for the model.

This formulation is slightly different than Jacob~\cite{jacob_learning_2024} and Munn~\cite{Munn_2026}, who instead match the deflection angle at the end of the beam and use $K=\frac{EI}{L}$. 

\subsection{Flexural modulus testing}
\label{subsec:modulustesting}
We gathered northern highbush blueberry cane samples from a research farm during the dormant season in Winter 2026 from the `Legacy', `EarliBlue', `Elliott', and `Duke' varieties, and during the fruiting season in Summer 2026 from the `Duke', `Liberty', and `Draper' varieties. We chose samples that were at least 12 inches long, reasonably straight, and with no branching elements within the sample length.   

\begin{figure}
    \centering
    \vspace{5pt}
    \includegraphics[width=\linewidth]{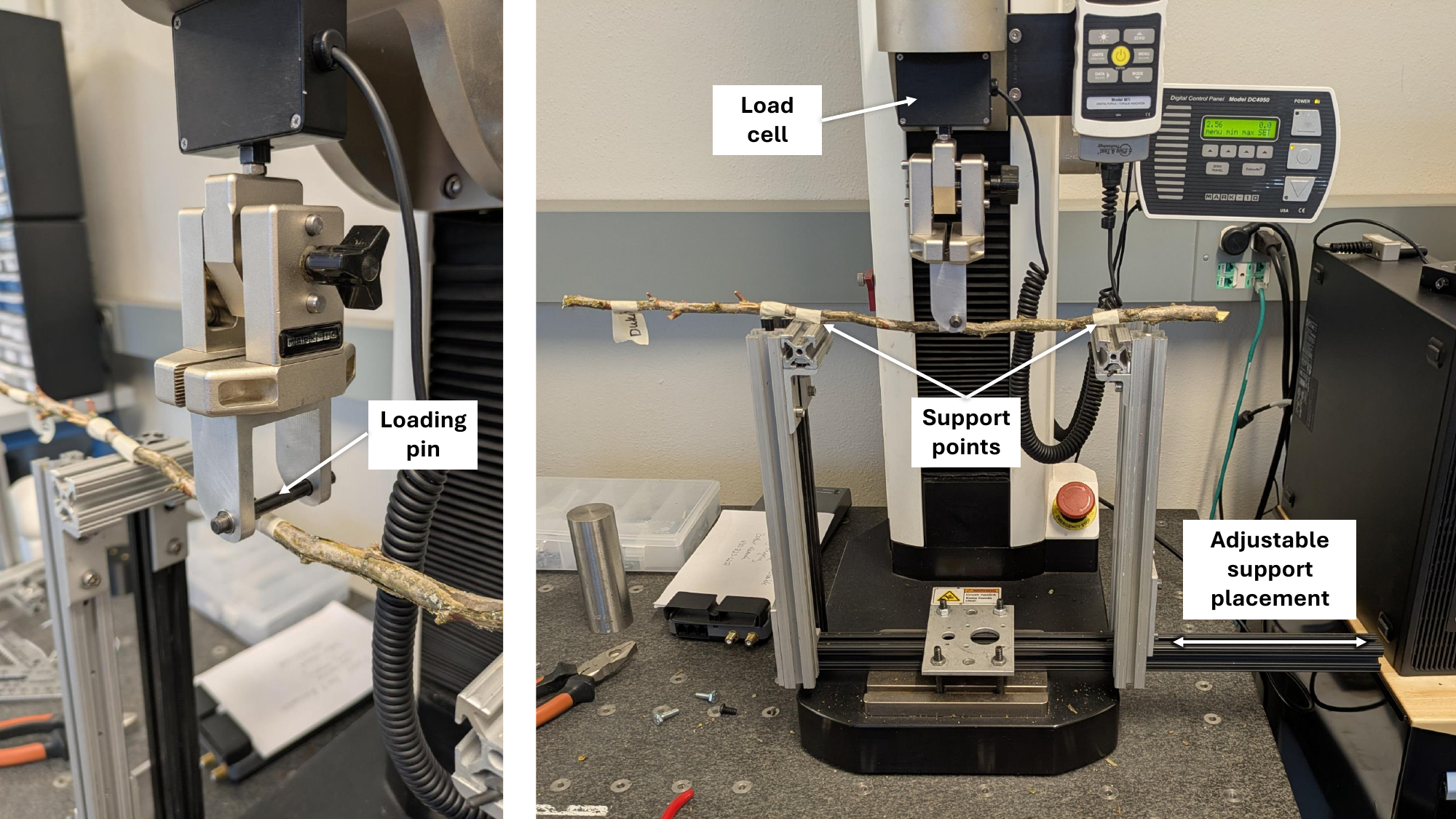}
    \caption{Fixture for 3-point flexural testing of blueberry cane samples in a Mark-10 tensile tester. \textit{Left:} Force was applied with a quarter-inch cylinder at the loading point.  \textit{Right:} The support width could be modified to accommodate various sample geometries.}
    \label{fig:mark10}
    \vspace{-12pt}
\end{figure}

We performed 3-point flexural testing within 30 hours of gathering the samples to capture the properties of ``green" wood as best as possible. Testing was performed with a Mark-10 tensile tester (Copiague, NY, USA) and a custom test fixture, as depicted in Fig.~\ref{fig:mark10}. The test fixture's support distance can be modified to accommodate different sample lengths. The ends of the sample were simply supported, and force was applied by a quarter-inch cylinder in the center. 

The support distance for each sample was chosen based on its particular geometry; 203 mm (8 in) was used as the default. We favored straight sections and avoided using knots or branch points as supports or the loading point. To run each test, the Mark-10 traveled at 50 mm/min downwards, exerting force on the sample until the sensor measured a predetermined force limit (between 20 and 30N), at which point data collection stopped. Then, the Mark-10 was manually stopped and jogged to its starting position. Each sample was tested three times with a 50N load cell. For each trial, we recorded a force-displacement curve. For samples that were linear until 20N was reached, the slope of the line was recorded. For samples that were not linear in that regime, such as thin, flexible new growth, a line of best fit was used to approximate the curve over the first 10mm of displacement at the default support width (\textit{Assumption \#~\themycounter}). The flexural modulus was then calculated as 
\stepcounter{mycounter}
\begin{equation}
    E_{flex} = \frac{L^3m}{12\pi r^4}
\end{equation}
where $E_{flex}$ is the flexural modulus (GPa), $L$ is the distance between the supports (mm), $m$ is the aforementioned slope (N/mm), and $r$ is the radius of the test sample (mm). Since the samples are not uniform in diameter across their length, we used an average of six caliper measurements to calculate the radius: the minimum and maximum diameters at both the support points and the loading point.

\subsection{Field Trial Contact Detection}
\label{sec:contact}
\begin{figure}
    \centering
    \includegraphics[width=\linewidth]{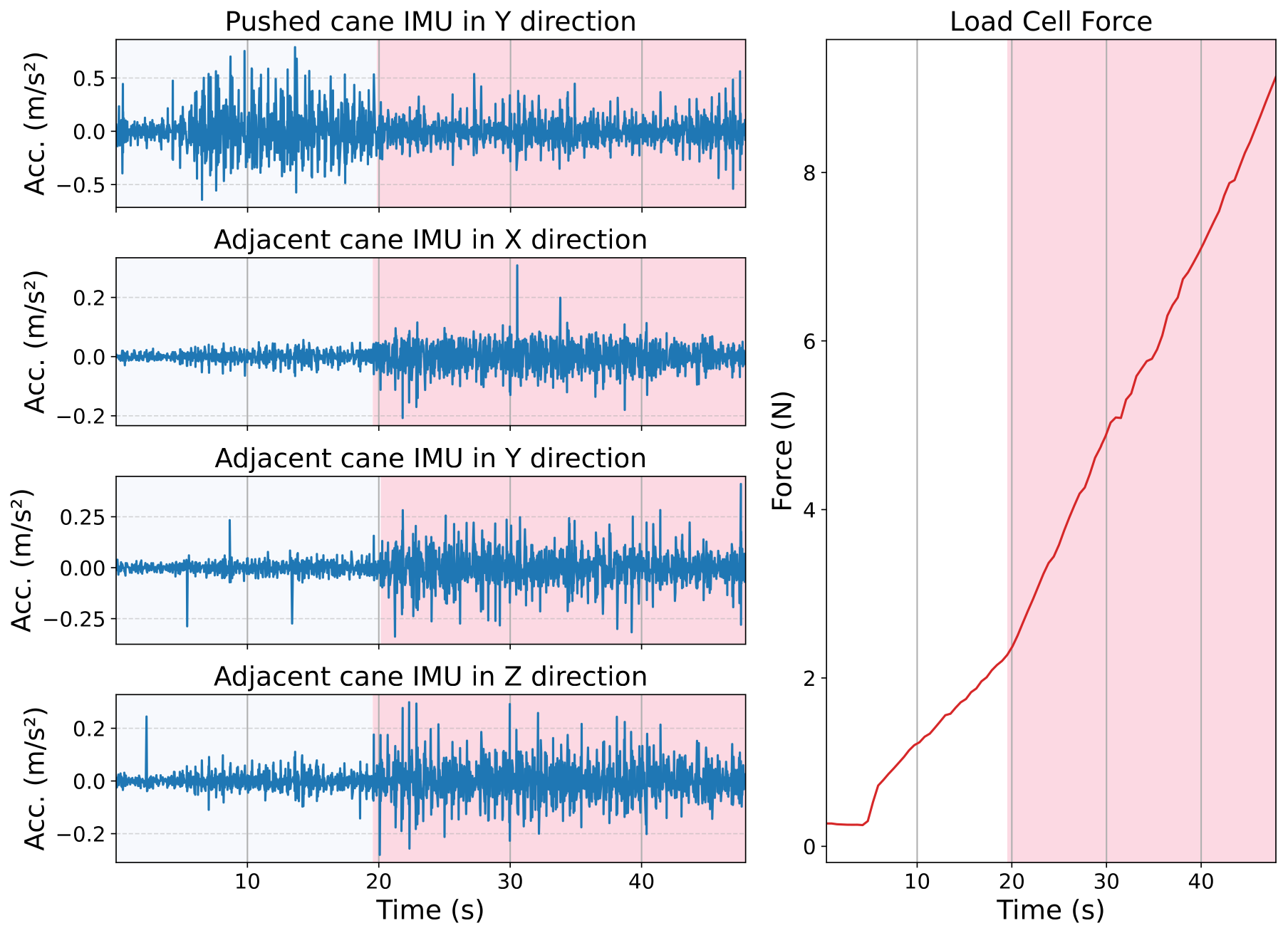}
    \caption{At the 20 second mark on this trial, the pushed cane hits an obstruction, likely the other cane. \textit{Left:} One accelerometer channel on the pushed cane and three accelerometer channels on an adjacent cane show a noise distribution change at the 20s mark, indicating contact. The change points are detected by the \textit{ruptures} library. \textit{Right:} The force-displacement graph shows an elbow at 20s, indicating that more force is suddenly required to push the cane, corroborating that contact occurred. This trial meets the criteria for contact, so the data is truncated at 20s during analysis.}
    \label{fig:imus}
    \vspace{-10pt}
\end{figure}

The purpose of the field data collected for this study is to evaluate a model of a single independent, unconstrained cane. Unfortunately, in several of the field trials the cane that was being measured was pushed into an obstruction, such as the supporting trellis structure or another cane. This contact was not detected on-site during data collection but is visible in the video. To address this --- and determine the actual contact point --- we analyzed the IMU data. Specifically, we rely on the acceleration measurements from the IMUs installed on the plant. In trials where no contact occurred, accelerometer noise was constant throughout probing, but when contact occurred, there was a distinct change in the accelerometer noise distribution, which coincided with an ``elbow" in the force-displacement plot (see Fig.~\ref{fig:imus} (\textit{right})). 

Accelerometer data was collected at 40Hz across three accelerometers: one on the pushed cane, and two on adjacent canes on the same bush, for nine total accelerometer channels. We first implemented a rolling average filter with a window size of 20 to account for the slow change in orientation during the push. To detect changes, we used the ruptures library \cite{Truong_Oudre_Vayatis_2020} to implement a Gaussian kernel-based detection of change points using binary segmentation with a penalty value of $22.5log(n)$, where $n$ is the length of the signal. Figure~\ref{fig:imus} shows the change points detected for a trial where contact occurred. 

Our criteria for contact was:
\begin{itemize}
\item Change-points are detected on 4 accelerometer channels, including one from the pushed branch, within a 1-second span, \textbf{or}
\item There exists an acceleration spike of greater than $1.5\frac{m}{s^2}$ on a pushed cane.
\end{itemize}

Of the 54 total trials, 12 met the criteria for contact. One of the twelve had so many spikes that we entirely excluded it, but the others we left as partial trials that ended as soon as contact was detected. One other trial push point was not visually captured by the camera data, leaving us with 41 full trials, 11 partial trials, and 2 exclusions.

%% file: Sections/04_Experiments.tex
\section{Experiments}
\label{sec:Experiments}
Our ultimate goal is to validate our rigid-body model against the real-world push data. First, we na\"{i}vely apply one ``best guess" set of modeling assumptions to every trial and compare the results to the experimental data on a trial-by-trial basis. We follow that up with a Method of Morris sensitivity analysis to examine which parameters are most influential on the model on a broader scale. 

To run a trial, a MuJoCo simulation is initialized with a cane as reconstructed by the method in Sec.~\ref{subsec:reconstruction} with all joints at their resting positions. To mimic the motion profile of the stepper motor used in the field with the testbed, the external force is modulated by a PID controller that increases its position setpoint by half a millimeter every second. Its parameters are $K_p = 750$, $K_i = 100$, and $K_d = 20$. The simulation and controller both iterate at 40 kHz. We use MuJoCo's implicit integrator and Newton solver. 

Each trial has a duration of 90 seconds or the duration of the field trial, whichever is shortest. All 52 full and partial trials are included; partial trials are truncated at contact. Figure~\ref{fig:pipeline}(\textit{d}) shows a MuJoCo simulation at the end of its run. Force and displacement data are recorded once at the end of each motor ``step" to allow oscillations to settle. We then compare the force-displacement data between the field trial and the simulation for equivalent perturbations on the cane. Figure~\ref{fig:pipeline}(\textit{e}) plots one trial's field data and simulation data over equivalent distances. 

\subsection{Na\"{i}ve case}
\label{subsec:OGTrial}
Canes were discretized as described in Fig.~\ref{fig:pipeline}. We used a flexural modulus of 4.57 GPa to calculate the joint stiffnesses for all joints in the model, the average of all the dormant season samples from Section \ref{subsec:modulustesting} (\textit{Assumption \#~\themycounter}). 

\subsection{Method of Morris}
\label{subsec:Morris}
We performed a Method of Morris sensitivity analysis over four variables. We set bounds for each based on its range of plausible values. 

\begin{enumerate}
\item The number of segments the cane is segmented into, set between 3 (the minimum that our discretization algorithm used) and 20 (after which we found that stiffness converges; \cite{Pivato_Dupont_Brunet_2014} suggests that there are few increasing gains over 10.)
\item The angle at which the force is applied to the cane, from -30 to 30 degrees from perpendicular, based on reasonable human error of setting the push bar. 
\item The flexural modulus, set to the minimum and maximum values measured from our testing, 1.82 to 8.05 GPa. 
\item The estimation of the diameter of the cane at unmeasured points. Using the centroid of the three measured points as an anchor point, we considered an upper bound of 1 to represent a linear fit of lengths and diameters, and a lower bound of 0 to represent no change of diameter along its length. Intermediate values change the slope of the relationship. 
\end{enumerate}

We used SALib \cite{Iwanaga2022, Herman2017} to run a 6-level sample with 4 trajectories on each of the 41 full and 11 partial trials.

%% file: Sections/05_Results.tex
\section{Results \& Discussion}
\label{sec:Results}

\subsection{Flexural modulus testing}
\label{subsec:flexmodresults}

\begin{figure}[b]
\centering
\includegraphics[width =\linewidth]{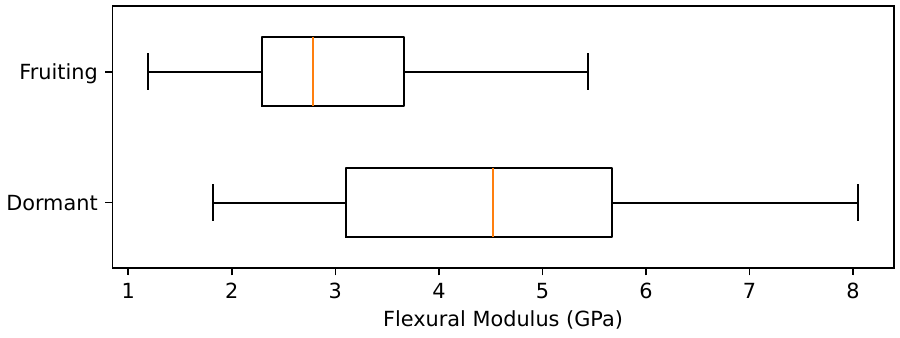}
\caption{The data shows a drop in the flexural modulus of northern highbush blueberry plants between the dormant season (\textit{n=19}) and the fruiting season (\textit{n=40})}
\label{fig:flexmodbyseason}
\vspace{-5pt}
\end{figure}

\begin{figure}[b!]
\centering
\includegraphics[width =\linewidth]{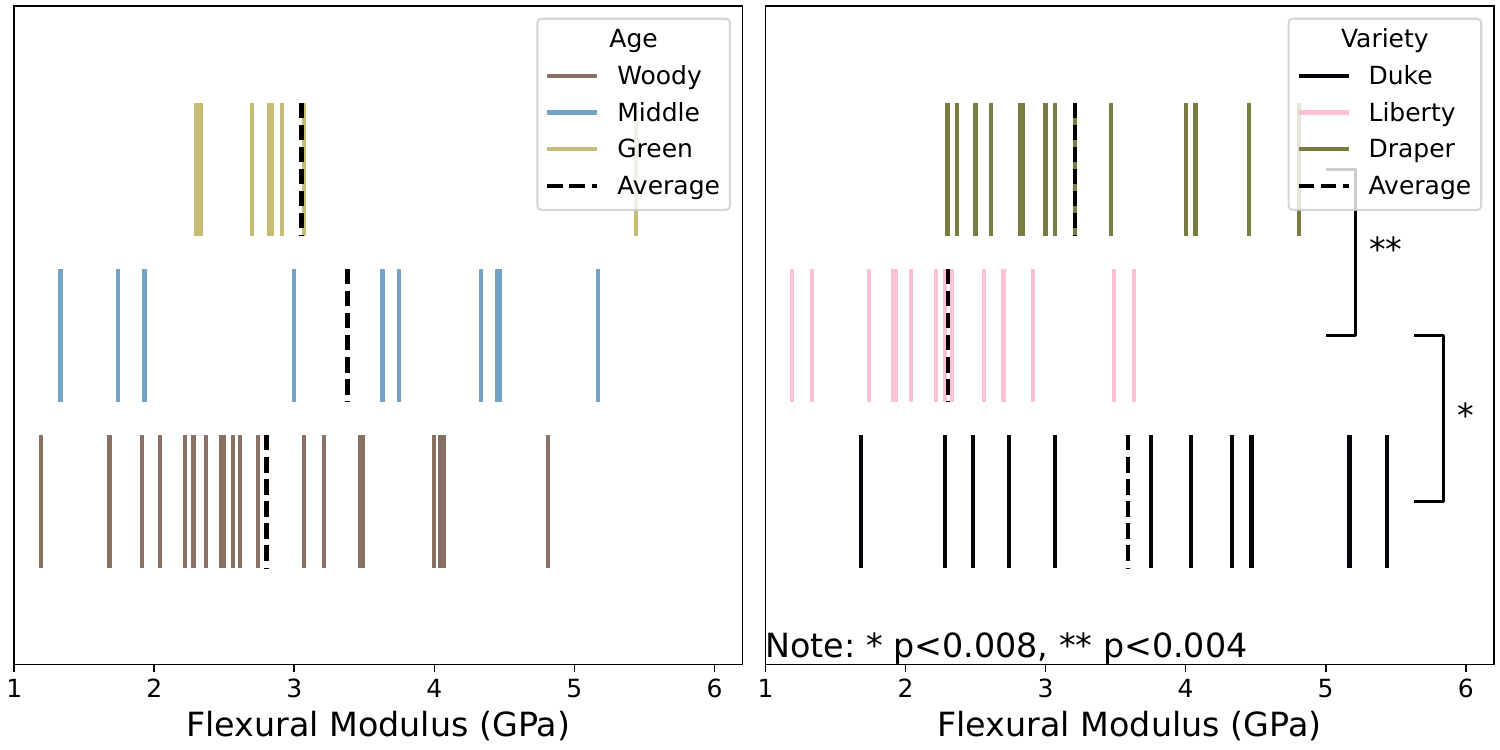}
\caption{Fruiting season flexural modulus measurements. Averages are indicated with black dashed lines. \textit{Left:} Fruiting data categorized by the age of the sample. \textit{Right:} Fruiting data categorized by the variety of the blueberry. Significant differences are marked via Welch's t-test.}
\label{fig:flexmodbyvariety}
\end{figure}

Figure~\ref{fig:flexmodbyseason} summarizes the seasonal decrease in modulus from the dormant season to the fruiting season. Even though we only had 19 total samples in the dormant season and 40 total samples in the fruiting season, a Welch's t-test yields a p-value of 0.0015, indicating a significant difference between the seasons. The mean is 4.57 GPa in the dormant season and 3.00 GPa in the fruiting season. Figure~\ref{fig:flexmodbyvariety} shows the distributions of measurements by age and variety for the fruiting season.  

With our calculation method, we found no statistically significant difference between samples of different ages during the fruiting season. However, there were significant differences between the pairwise means of Duke and Liberty samples, as well as Draper and Liberty samples. 

\subsubsection*{Discussion}
We recognize that our evidence would be stronger if the same varieties of blueberries were tested in both the fruiting and dormant seasons. However, the Duke variety did have overlap, with  5 samples in the dormant season and 11 samples in the fruiting season. Duke samples matched the overall trend, with an average flexural modulus of 4.33 GPa in the dormant season and 3.59 GPa in the fruiting season, but did not have statistical significance on its own due to the small sample size. Therefore, we can only lightly conclude the downward trend from dormant season to fruiting season because of the confounding factor of the variety. However, our data does corroborate that of Cannell and Morgan, which suggests that flexural modulus is correlated with water content, which is higher in blueberries during the fruiting season [17].
 
Because we found a statistically significant difference between particular varieties during the fruiting season (Fig.~\ref{fig:flexmodbyvariety} (\textit{right})), a flexural modulus should be found for each variety individually. However, since we did not collect enough samples from the Legacy variety in the dormant season, which is what our experimental data was collected on, we had to rely on an average for our ``na\"{i}ve" experiment. Additionally, the variation between samples was large, which could be caused by any number of factors: uneven diameters along the length caused by buds, spurs, smaller branches, or uneven growth; variation in the measurement of that diameter; or actual variation between samples, which was pointed out by [16]. In fact, it may be impossible to specify the modulus of a branch before making contact.

It should also be pointed out that ``green" canes exhibited significantly more plastic properties than ``woody" canes. As described in Sec.~\ref{subsec:modulustesting}, we made a linear approximation for small deformations, so extrapolation to large deformations may be inaccurate. 

\subsection{Na\"{i}ve case}

\begin{figure*}[]
\centering
\vspace{5pt}
\includegraphics[width =\linewidth]{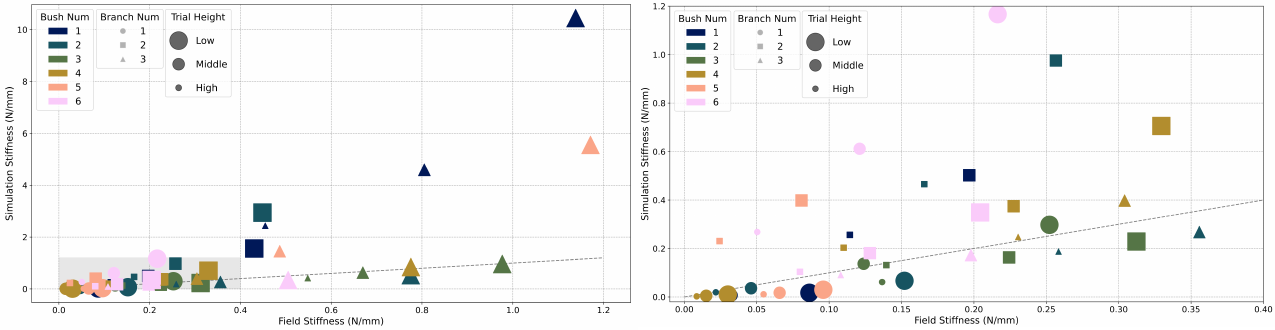}
\caption{A comparison of effective stiffness values on a per-trial basis between the real-world data and the simulation. \textit{Left:} The full plot of stiffness values plotted against a 1:1 ideal line. Trials from the same bush have the same marker color; the shape of the marker indicates the branch number, and the size indicates the height of the probing point (Low, Middle, High). Trials above the line indicate that the simulation has a stiffness value higher than the real data; trials under the line indicate the simulation was too compliant.   \textit{Right:} An enlarged plot of the gray boxed area from \textit{left}.}
\label{fig:doubleplot}
\vspace{-10pt}
\end{figure*}

For each of the 52 full and partial trials, we calculated the slope of the force-displacement data for the field trial and the simulation run (Fig.~\ref{fig:pipeline}(\textit{e})). Figure~\ref{fig:doubleplot} compares those slopes on a per-trial basis. Stiffness varied in the range of 0.009 to 1.2 N/mm in the field, and 0.003 to 10.4 N/mm in simulation. Figure~\ref{fig:logplot} (\textit{top}) shows the log-ratio errors between the simulation and field data.

\subsubsection*{Discussion} 
The best alignment between simulation and trial data is almost always the highest elevation trial location on the cane, followed by the middle, and then the lowest. One possible explanation is a misrepresentation of the diameters at low heights. Since the trials lowest to the ground are the ones most reliant on extrapolated data (the diameters of the links below them are all estimated), it stands to reason that they are the most affected by the method of extrapolation. Other models of height/diameter relationships, such as one of the many cited by Baia et al.~\cite{Baia_Nascimento_Guedes_Hilario_Toledo_2025}, may be better suited to blueberry canes than the linear assumption and could potentially lead to better results. Alternatively, trunk or branch diameter estimation methods via machine vision could be used as a more automated method of diameter extraction~\cite{Ahmed_Sapkota_Churuvija_Karkee_2025, Brown_Paudel_Biehler_Thompson_Karkee_Grimm_Davidson_2024}. Regardless, a `local' diameter is insufficient for estimating the force characteristics of a cane; a broader reconstruction is necessary. For blueberry canes where the base is dense and occluded with many shoots, skeletonization and diameter estimation will be a nontrivial task and is left to future work. 

Other potential sources of error are flexions in the testbed system; the use of an average dormant season flexural modulus instead of a variety-specific flexural modulus; or other modeling parameter inaccuracies such as damping or density. We note some of the most egregious errors, such as on Bush 1, Branch 3, almost met the contact exclusion criteria, but were ultimately not truncated, so we believe there could be a strong correlation between high errors and missed contacts, and a better method for identifying contact should be identified. 

\subsection{Method of Morris sensitivity analysis}

\begin{figure}
\centering
\includegraphics[width =\linewidth]{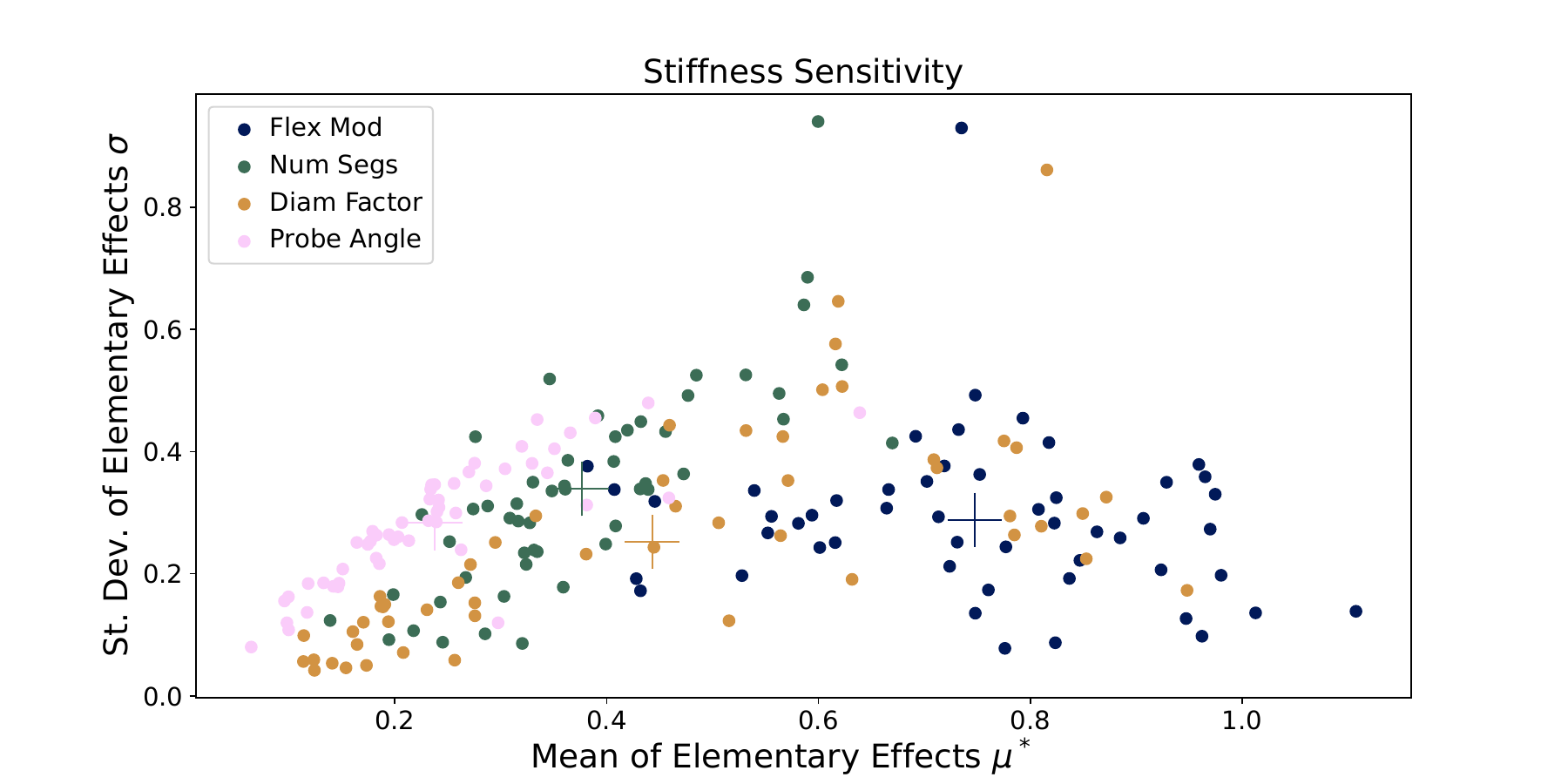}
\caption{Output visualization of the Method of Morris study. Each dot is one study performed on one trial; the cross represents the mean over all simulation trials.}
\label{fig:morrisplot}
\vspace{-10pt}
\end{figure}

We performed sensitivity analysis over the raw stiffness of the simulated cane. Since the expected values were different for each trial, a separate Method of Morris study was conducted for each. The results are plotted on Fig.~\ref{fig:morrisplot}. 


As expected, the estimated flexural modulus was the most important factor. The probe angle had low $\mu^*$ values, indicating that it was not as important to the stiffness, or contributed as much to the error. The diameter factor and number of segments landed in the middle. Whereas we do see a handful of outliers, no variables grossly exceeded a 1:1 mean to standard deviation line, so there were no significant coupling or nonlinear effects between variables. 


\subsubsection*{Discussion}
The second most important factor appears to be the diameter estimation factor; yet, in some trials, this stand-in factor was just as important as the flexural modulus, and in some, it didn't matter at all. We suspect that this bifurcation is an artifact of the mathematical construction of the variable: if the three measurements of the cane suggested a uniform-diameter cane, then the sweep from a uniform-diameter cane to a cane defined by a linear fit through those measurements would not have a large effect on the output. However, for canes where the three measurements suggested a steep change in the diameter based on its length, the difference would be dramatic. 

Whereas the particular construction has its flaws, the study still makes an important point. In trials where the diameter of the cane at the base does change, that change is significant to the overall stiffness of the system, and the flexural modulus, which has a high sample-by-sample variance, outweighs the other factors significantly.

\subsection{Parameter Sweep}
\begin{figure}
\centering
\includegraphics[width =\linewidth]{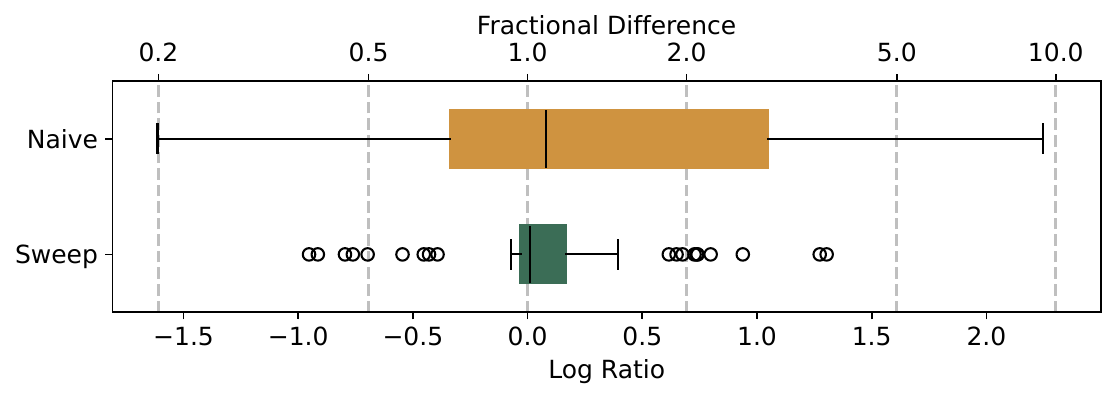}
\caption{The log-ratio error of each trial on the Na\"{i}ve case and the lowest log-ratio error found when sweeping flexural modulus and diameter factor.}
\label{fig:logplot}
\vspace{-10pt}
\end{figure}

Lastly, as a final check of whether the model is suitable for modeling blueberry canes, we performed a parameter sweep on the two most influential variables (flexural modulus and diameter factor) at five uniform levels with the same bounds as the Morris test. We recorded the lowest log-ratio error found on that sweep in Fig.~\ref{fig:logplot}. 

\subsubsection*{Discussion} Whereas choosing one value for each of the modeling factors produces errors as high as 10x the measured values, a sweep over the two most influential factors shows that, in most cases, \textbf{a reasonable force-displacement model does lie in the distribution space of the modeling parameters, but it is not the same model for every cane.} Therefore, future modelers should not na\"{i}vely choose singular values like flexural modulus when using such models to learn policies for interaction, and rather pull from  distributions in order to cover the natural variations of plants.

%% file: Sections/06_Conclusion.tex
\section{Conclusion}
\label{sec:Conclusion}
We would first like to recognize limitations on our conclusions based on the methods we used. 
\begin{itemize}
    \item \textbf{Simple canes:} Our tests were limited to simple canes without any branching structures, so we cannot make conclusions about whether the lumped-parameter method extends to more complex geometries. However, we believe the simple case was the best place to start in regards to understanding the underlying mechanisms. 
    \item \textbf{Small deflections:} Based on the short distance traveled and the small angle assumption, our results are only valid for small deflections, and may not extrapolate to large deflections, i.e. moving a cane to the ground. 
    \item \textbf{Sample size:} Due to time limitations on the commercial farm, our sample size is small. However, we do believe it is commensurate with other agricultural robotics test sizes. 
\end{itemize}

Despite these limitations, our study still holds several insights. From the Method of Morris sensitivity analysis, it is clear that having a well-tuned flexural modulus is the most important modeling factor to drive down the error between the experimental and simulated plant behavior under loading. Unfortunately, the flexural modulus of a plant is difficult to determine, as it changes with a variety of factors. \textbf{Simulators intent on accuracy should make sure to consider both the variety and the season; machine learning models should train on a distribution of plant stiffnesses in order to be robust to seasonal, varietal, and individual variations.} 

Additionally, accurate phenotyping of the plant (i.e. geometric reconstruction, particularly the diameter) is both a technical challenge and particularly important to deformation behavior. \textbf{A ``local" reconstruction is insufficient; a good digital model will require accurate reconstruction down to the base. }

Overall, we believe the lumped-parameter approach to be a reasonable model for deformable plants, in particular for applications regarding small deformations, such as mass loading during fruit maturation or pruning tasks, but cannot confirm its usability for large deflections. 


\section*{Acknowledgments}
Claude Sonnet was used to generate code for the first drafts of several plots and for debugging instable simulation instances. All plots and figures included in the paper were produced by the authors. 